\documentclass[11pt]{article}

\usepackage[final]{style/acl}

\usepackage{times}
\usepackage{latexsym}
\usepackage{microtype}
\usepackage{hyperref}
\usepackage{multirow}
\usepackage{float}
\usepackage{graphicx}
\usepackage{tabularx}
\usepackage{svg}
\usepackage[most]{tcolorbox}
\usepackage{cleveref}
\crefname{section}{§}{§§}
\usepackage[T1]{fontenc}
\usepackage{inconsolata}
\usepackage{t1enc}
\usepackage{enumitem}
\usepackage{longtable}
\usepackage{url}
\usepackage{booktabs}
\usepackage{makecell}
\usepackage{lineno}

\usepackage{framed}
\usepackage{mdwlist}
\usepackage{siunitx}
\usepackage{latexsym}
\usepackage{colortbl}
\usepackage{xcolor}
\usepackage{nicefrac}
\usepackage{booktabs}
\usepackage{fnpct}
\usepackage{amsfonts}
\usepackage[T1]{fontenc}
\usepackage{bold-extra}
\usepackage{amsmath}
\usepackage{amssymb}
\usepackage{bm}
\usepackage{graphicx}
\usepackage{mathtools}
\usepackage{microtype}
\usepackage{multirow}
\usepackage{multicol}
\usepackage{todonotes}
\usepackage{xpatch}
\usepackage{latexsym,comment}
\usepackage[normalem]{ulem}

\newcommand*{\missingreference}{{\Huge \colorbox{red}{?reference?}}}
\newcommand*{\missingcitation}{{\Huge \colorbox{red}{?citation?}}}

\makeatletter
\xpatchcmd{\@setref}{\bfseries}{\missingreference}{}{}
\def\@citex[#1]#2{\leavevmode
    \let\@citea\@empty
    \@cite{\@for\@citeb:=#2\do
        {\@citea\def\@citea{,\penalty\@m\ }%
            \edef\@citeb{\expandafter\@firstofone\@citeb\@empty}%
            \if@filesw\immediate\write\@auxout{\string\citation{\@citeb}}\fi
            \@ifundefined{b@\@citeb}{\hbox{\reset@font\missingcitation}%
                \G@refundefinedtrue
                \@latex@warning
                {Citation `\@citeb' on page \thepage \space undefined}}%
            {\@cite@ofmt{\csname b@\@citeb\endcsname}}}}{#1}}
\makeatother

\newcommand{\gem}[1]{\mbox{\textsc{gem}}}
\newcommand{\abr}[1]{\textsc{#1}}

\newcommand{\hidetext}[1]{}
\newcommand{\ignore}[1]{}

\newcommand{\specialcell}[2][c]{%
  \begin{tabular}[#1]{@{}c@{}}#2\end{tabular}}

\newcommand{\specialcellleft}[2][l]{%
\begin{tabular}[#1]{@{}l@{}}#2\end{tabular}}

\newcommand{\smallurl}[1]{ \begin{tiny}\url{#1}\end{tiny}}

\definecolor{lightblue}{HTML}{3cc7ea}
\definecolor{CUgold}{HTML}{CFB87C}
\definecolor{grey}{rgb}{0.95,0.95,0.95}
\definecolor{ceil}{rgb}{0.57, 0.63, 0.81}
\definecolor{UMDred}{HTML}{ed1c24}
\definecolor{UMDyellow}{HTML}{ffc20e}
\definecolor{darkblue}{rgb}{0, 0, 0.5}
\hypersetup{colorlinks=true, citecolor=darkblue, linkcolor=darkblue, urlcolor=darkblue}
\definecolor{purple_plot}{HTML}{800080}
\definecolor{green_plot}{HTML}{008000}

\usepackage[T1]{fontenc}

\usepackage[utf8]{inputenc}

\usepackage{microtype}

\usepackage{inconsolata}

\usepackage{graphicx}

\title{Arguments that Alter Minds: LLM Rationales Sway \\Human (and LLM) Notions of Plausibility}

\newcommand{\authorsep}{\hspace*{10mm} \quad}

\author{Shramay Palta\authorsep Peter Rankel\\
\bf Sarah Wiegreffe\authorsep Rachel Rudinger\\
  University of Maryland, College Park \\ 
  \texttt{\{spalta,par,sarahwie,rudinger\}@umd.edu}
  }

\newcommand{\colorBoxColorPro}{purple_plot}
\newcommand{\colorBoxColorCon}{UMDred}
\newcommand{\colorBoxColor}{UMDyellow}
\newcommand{\colorBoxColorProCon}{green_plot}
\begin{document}
\newtcolorbox[list inside=prompt,auto counter,number within=section]{prompt}[1][]{
    colbacktitle=black!60,
    fonttitle=\small,
    coltitle=white,
    fontupper=\footnotesize,
    boxsep=3pt,
    left=0pt,
    right=0pt,
    top=0pt,
    bottom=0pt,
    boxrule=1pt,
    width=\linewidth, 
    enlarge left by=0mm, 
    enlarge right by=0mm, 
    #1
}

\maketitle
\begin{abstract}
We investigate the degree to which human (and \textsc{LLM}) plausibility judgments of multiple-choice commonsense benchmark answers are subject to influence by (im)plausibility arguments for or against an answer, in particular, using rationales generated by \textsc{LLM}s. We collect $3,000$ plausibility judgments from humans and another $13,600$ judgments from \textsc{LLM}s. Overall, we observe increases and decreases in mean human plausibility ratings in the presence of \textsc{LLM}-generated \texttt{PRO} and \texttt{CON} rationales, respectively, suggesting that, on the whole, human judges find these rationales convincing. Experiments with \textsc{LLM}s reveal similar patterns of influence. Our findings demonstrate a novel use of \textsc{LLM}s for studying aspects of human cognition, while also raising practical concerns that, even in domains where humans are ``experts'' (i.e., common sense), \textsc{LLM}s have the potential to exert considerable influence on people's beliefs.\footnote{Our data is available at \url{https://github.com/shramay-palta/AAM.git}.}
\end{abstract}
\section{Introduction}
A typical framework for evaluating the commonsense reasoning abilities of \textsc{NLP} models is to present an everyday situation described in text (\textit{A person drops a glass}), followed by a set of possible inferences (\textit{The glass breaks}; \textit{The glass bounces}; etc), and instruction to select the most plausible answer choice~\citep{Levesque2011TheWS, roemmele2011choice, mostafazadeh-etal-2016-corpus}. Unlike in logical or mathematical reasoning problems, these answer choices are rarely definitively true or false, but rather lie on a continuum of plausibility \citep{zhang-etal-2017-ordinal, chen-etal-2020-uncertain}, and require non-monotonic modes of reasoning to assess, such as defeasible \citep{rudinger-etal-2020-thinking}, abductive \citep{Bhagavatula2020Abductive}, or counterfactual \citep{qin-etal-2020-back} reasoning. In this setting, the answer designated as ``correct'' should be the choice that human judges agree is most plausible. Indeed, human annotators are more likely to agree on a ``correct'' answer when the plausibility gap between answer choices is large \citep{palta-etal-2024-plausibly}.

However, unless an answer choice is logically impossible, there exist circumstances under which even a relatively implausible distractor answer could be true. For example, one could argue that a glass is actually quite likely to bounce if dropped on a trampoline or rubber mat. By the same token, unless an answer choice is logically necessary, there exist circumstances under which an otherwise highly plausible outcome becomes much less likely: The drinking glass probably \textit{won’t} break if it’s very sturdy or doesn’t fall far. Thus, any answer choice which is neither certain nor impossible can be subject to arguments for or against its plausibility. In principle, these types of (im)plausibility arguments or rationales do not introduce any new evidence to the situation; rather, they simply highlight possible circumstances which, \textit{if true}, would impact an answer’s plausibility.

\begin{figure*}[t!]
    \centering   
    \includegraphics[width=\linewidth]{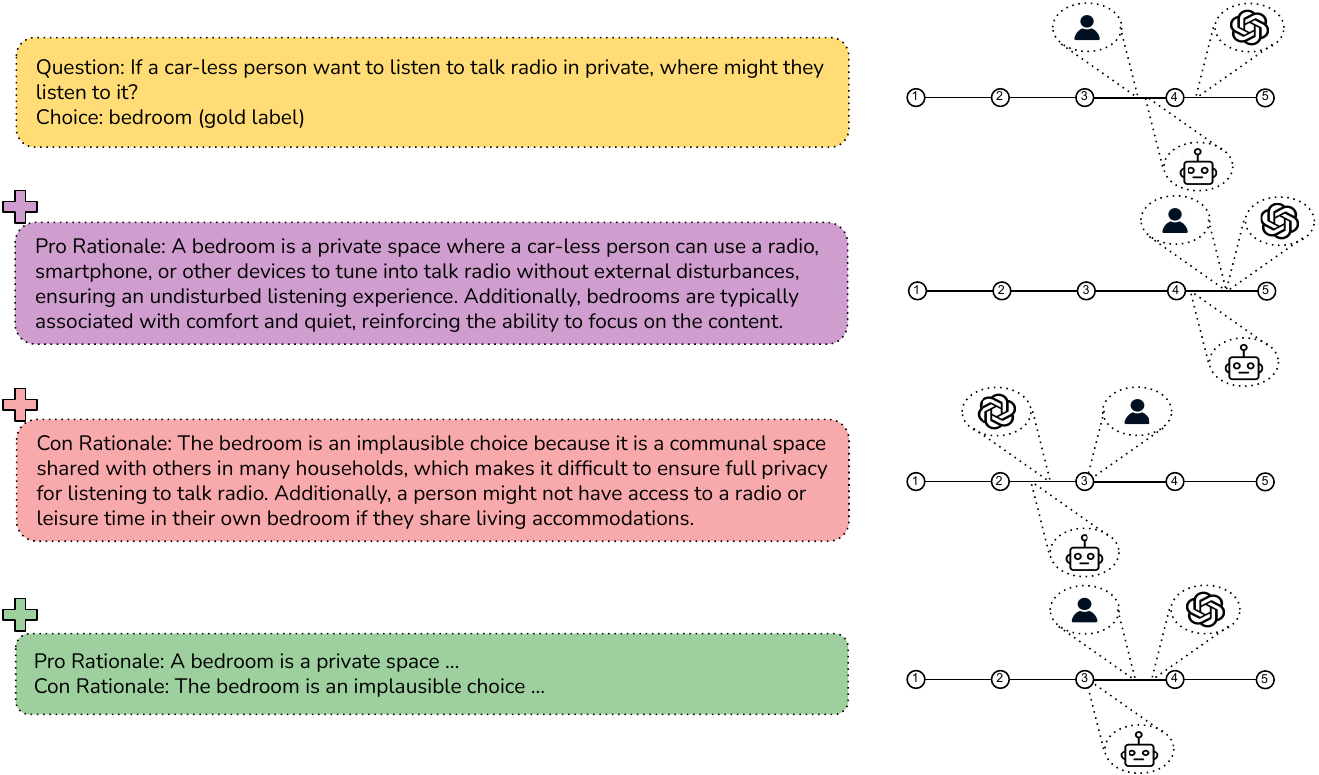}
    \caption{An example question and answer choice from \textsc{CQA} \citep{talmor-etal-2019-commonsenseqa} in \textcolor{UMDyellow}{yellow}, paired with \textsc{LLM}-generated rationales in favor of the answer's plausibility (\texttt{PRO}, \textcolor{purple_plot}{purple}); against the answer choice's plausibility (\texttt{CON}, \textcolor{UMDred}{red}); and both for and against (\texttt{PRO}+\texttt{CON}, \textcolor{green_plot}{green}). Likert scales on right show how mean plausibility ratings for the answer, as judged by human and \textsc{LLM} raters (OpenAI and Non-OpenAI), shift in response to different types of rationales.}
    \label{fig:example}
\end{figure*}

In this paper, we investigate whether an \textsc{LLM} can produce convincing arguments for the plausibility or implausibility of commonsense benchmark answer choices, as measured by their ability to impact the plausibility ratings of human (and LLM) judges. (We will refer to such plausibility or implausibility arguments as \texttt{PRO} or \texttt{CON} rationales, respectively.) Posed another way, this paper asks whether and how human plausibility judgments change in response to LLM-written \texttt{PRO} and/or \texttt{CON} rationales.

To answer these questions, our work builds on the prior study of \citet{palta-etal-2024-plausibly}, who previously collected Likert-scale judgments from human annotators about the plausibility of individual question-answer choice pairs from two commonsense multiple choice benchmarks, Social IQA (\textsc{SIQA}; \citealt{sap-etal-2019-social}) and CommonsenseQA (\textsc{CQA}; \citealt{talmor-etal-2019-commonsenseqa}). These judgments were collected in the absence of any rationales or (im)plausibility arguments, which we will refer to as the \texttt{NO} rationale setting. We extend this work by sampling a subset of their annotated question-answer pairs, and using a \textsc{LLM} to generate \texttt{PRO} and \texttt{CON} rationales for or against the plausibility of each answer. Adopting a similar protocol, we then collect $3,000$ new human plausibility ratings of answer choices, presented alongside either a \texttt{PRO} rationale of the answer, a \texttt{CON} rationale, or both (\texttt{PRO+CON}), as illustrated in Figure~\ref{fig:example}. In a parallel set of experiments using the same stimuli, we prompt $17$ different \textsc{LLM}s to also rate the plausibility of these answers, collecting an additional $13,600$ judgments from \textsc{LLM}s.

\begin{table*}[!t]
\centering
\small
\begin{tabular}{c|cccccc}
\toprule
Dataset & \multicolumn{1}{c|}{\#$q$} & \multicolumn{1}{c|}{\begin{tabular}[c]{@{}l@{}}\#$(q,a)$ \\\end{tabular}} & \multicolumn{1}{c|}{\begin{tabular}[c]{@{}l@{}}\#$(q,a,r)$\end{tabular}} & \multicolumn{1}{c|}{\begin{tabular}[c]{@{}c@{}}\#Human\\Judgments \end{tabular}} & \multicolumn{1}{c|}{\begin{tabular}[c]{@{}c@{}} \#LLM\\Judgments\end{tabular}} & \begin{tabular}[c]{@{}c@{}}\#Total\\Judgments\end{tabular} \\ \midrule
\textsc{SIQA} & $50$ & $100$ & $400 $ & $1,500(+500)$ & $6,800 $ & $8,300(+500)$ \\ \midrule
\textsc{CQA} & $50$ & $100$ & $400 $ & $1,500(+500)$ & $6,800$ & $8,300(+500)$ \\ \midrule
\textbf{Total} & 100 & 200 & 800 & $\textbf{3,000(+1,000)}$ & $\textbf{13,600}$ & $\textbf{16,600(+1,000)}$ \\ \bottomrule
\end{tabular}
\caption{Size and structure of the collected dataset. From left to right, columns report number of unique questions drawn from each dataset (\#$q$); number of unique question-answer pairs (\#$(q,a)$); number of unique question-answer-rationale tuples (\#$(q,a,r)$) over four settings (\texttt{NO} rationale, \texttt{PRO} rationale, \texttt{CON} rationale, and \texttt{PRO+CON} rationale); total number of human plausibility ratings collected; total number of \textsc{LLM} plausibility ratings generated; and total number of plausibility ratings (human or \textsc{LLM}). Numbers in parentheses represent \textit{additional} judgments previously collected by \citet{palta-etal-2024-plausibly} for the \texttt{NO} rationale setting, used in our analysis.}
\label{tab:anno_stats}
\end{table*}

With this data, we are able to make the following observations:
\begin{itemize}[nosep]
    \item Both human and \textsc{LLM} plausibility judgments are significantly affected by the addition of \texttt{PRO, CON}, and \texttt{PRO+CON} rationales, relative to a \texttt{NO} rationale setting.
    \item On average, \texttt{PRO} rationales raise mean plausibility ratings from humans and \textsc{LLM} judges alike, while \texttt{CON} rationales lower mean ratings.
    \item These patterns hold across both gold and distractor answer choices, with one notable exception: mean plausibility ratings by humans surprisingly \textit{drop} for gold answer choices when a \texttt{PRO} rationale is presented. For \textsc{LLM}s, mean plausibility ratings increase in response to \texttt{PRO} rationales, for gold and distractor answer choices alike.
    \item \texttt{PRO+CON} rationales yield mixed reactions from both humans and \textsc{LLM}s, with ratings decreasing for the gold label, but increasing for the distractor answer choice. 
    \item While OpenAI and Non-OpenAI models generally exhibit similar trends, we observe some differences. In particular, \texttt{CON} rationales appear to have a larger effect on the OpenAI models' ratings compared to the human ratings and Non-OpenAI models' ratings.
    \item The initial plausibility rating of an answer choice has a strong anchoring effect: the higher the initial plausibility of an answer, the smaller the subsequent change in its rating after a rationale is provided.
\end{itemize}
\section{Rationale Generation}\label{rationale_generation}
\citet{palta-etal-2024-plausibly} introduced a plausibility rating framework for rating the plausibility of individual answer choices from Multiple Choice Question (\textsc{MCQ}) datasets in isolation. For our study, we randomly sample a subset of 100 questions-- 50 each from \textsc{SIQA} and \textsc{CQA}-- from the set of questions used in \citet{palta-etal-2024-plausibly}. \textsc{SIQA} is designed to evaluate the social and emotional intelligence in computational models. Each \textsc{SIQA} item consists of a short context describing a social situation, a question about a person in the situation and three answer choices. \textsc{CQA}, on the other hand, evaluates general commonsense knowledge, with each item containing five answer choices. For each sampled question, we pick the dataset gold label ($a_{gold-label}$) and a randomly selected distractor answer choice ($a_{distractor}$) for our study. To maintain consistency with \citet{palta-etal-2024-plausibly}, we break down each question $q$ with answer choices $a_{gold-label}$ and $a_{distractor}$ into pairs $(q, a)$ where $a \in [a_{gold-label}, a_{distractor}]$.

We then generate two forms of rationales for each $(q, a)$ pair:
\begin{enumerate}[nosep]
    \item \textbf{\texttt{PRO} Rationale}: An explanation that best explains the \textit{plausibility} of the answer choice for the given question and context (for \textsc{SIQA}).
    \item \textbf{\texttt{CON} Rationale}: An explanation that best explains the \textit{implausibility} of the answer choice for the given question and context (for \textsc{SIQA}).
\end{enumerate}

These two rationales are then used to collect human and \textsc{LLM} plausibility judgments under four settings, described in \S\ref{sec:human_ratings}: \texttt{NO}, \texttt{PRO}, \texttt{CON}, and \texttt{PRO+CON}.
\begin{table*}[t]
\centering
\resizebox{\textwidth}{!}{
\begin{tabular}{|c|c|ccc|ccc|ccc|}
\toprule
\multirow{2}{*}{\textbf{Dataset}} & \multirow{2}{*}{\textbf{Agent}} 
& \multicolumn{3}{c|}{\textbf{Pro Rationale}} 
& \multicolumn{3}{c|}{\textbf{Con Rationale}} 
& \multicolumn{3}{c|}{\textbf{Pro+Con Rationales}} \\  
& & Increase & Decrease & No Change 
  & Increase & Decrease & No Change 
  & Increase & Decrease & No Change \\ 
\midrule
\multirow{3}{*}{SIQA} 
  & Human & $28\%$ & $22\%$ & $50\%$ & $2\%$ & $69\%$ & $29\%$ & $11\%$ & $44\%$ & $45\%$ \\ \cmidrule{2-11}
  & OpenAI & $40\%$ & $2\%$ & $58\%$ & $0\%$ & $84\%$ & $16\%$ & $12\%$ & $43\%$ & $45\%$ \\ \cmidrule{2-11}
  & Non-OpenAI & $61\%$ & $1\%$ & $38\%$ & $5\%$ & $66\%$ & $29\%$ & $20\%$ & $24\%$ & $56\%$ \\ \midrule
\multirow{3}{*}{CQA}  
  & Human & $30\%$ & $27\%$ & $43\%$ & $9\%$ & $33\%$ & $58\%$ & $13\%$ & $44\%$ & $43\%$ \\ \cmidrule{2-11}
  & OpenAI & $62\%$ & $4\%$ & $34\%$ & $4\%$ & $74\%$ & $22\%$ & $21\%$ & $38\%$ & $41\%$ \\ \cmidrule{2-11}
  & Non-OpenAI & $53\%$ & $5\%$ & $42\%$ & $4\%$ & $65\%$ & $31\%$ & $21\%$ & $32\%$ & $47\%$ \\ \bottomrule
\end{tabular}
}
\caption{Percentage cases where the mean plausibility rating given by human annotators, OpenAI and Non-OpenAI models increased, decreased or remained the same when different rationales were presented for \textsc{SIQA} and \textsc{CQA}. Numbers here include both $a_{gold-label}$ and $a_{distractor}$ answer types.}
\label{tab:rating_changes_agents}
\end{table*}
\subsection{Model Preference Study}\label{sec:preference}
Because the cost of human annotation is high (\$1,000 USD for our full-scale study), we first ran a small-scale preference study to select the language model most likely to produce rationales convincing to humans for subsequent use in the full-scale study. 

For this initial preference study, we prompted several state-of-the-art models: \texttt{GPT-4o} and \texttt{GPT-4o-mini} \citep{openai2024gpt4technicalreport}, as well as \texttt{Llama 3.1-8B-Instruct} and \texttt{Llama 3.1-70B-Instruct} \citep{grattafiori2024llama3herdmodels} to generate both \texttt{PRO} and \texttt{CON} rationales for a set of $40$ randomly sampled $(q, a)$ pairs from the \texttt{validation} split of \textsc{SIQA}. The prompts used for generating the \texttt{PRO} and \texttt{CON} rationales are presented in Prompts~\ref{prompt:rationale_generation_pro} and~\ref{prompt:rationale_generation_con} in Appendix~\ref{prompts}.

The generated rationales, and their corresponding $(q, a)$ pairs, were presented to $4$ human annotators, who selected the rationale ``that best explained the plausibility (for \texttt{PRO}) or implausibility (for \texttt{CON}) of the answer choice $a$, and was also the most believable''. We show an example of the interface presented to the annotators in \autoref{preference_interface}. Based on the annotators' responses, \texttt{GPT-4o} received the most votes and was selected to generate the rationales in our study. \footnote{According to annotator feedback, selecting a single rationale from the four presented was challenging, as the generated rationales were quite similar to one another. Nevertheless, \texttt{GPT-4o} received the majority of votes.} The set of $(q,a)$ pairs used in the preference study is entirely disjoint from the set used in the full-scale study.

We note that it is not the purpose of the preference study, nor the full-scale study, to determine \textit{which} \textsc{LLM} is \textit{most} persuasive to humans. Rather, the goal of the full-scale study is to determine whether \textit{any} \textsc{LLM} might produce rationales capable of swaying human plausibility judgment. A positive finding (that human judgments are indeed systematically swayed) can in principle be demonstrated with a single model, since it is not our claim that \textit{all} \textsc{LLM}s would necessarily yield such results. On the other hand, a negative finding (that the \textsc{LLM}-generated rationales did not affect human judgments) would beg the question: could a stronger model have yielded positive results? It is this latter concern that motivates our preference study.
\section{Human Plausibility Ratings}\label{sec:human_ratings}
To understand the impact of rationales on human (and LLM) plausibility judgments, we borrow the Likert scale introduced by \citet{zhang-etal-2017-ordinal}, and also used in \citet{palta-etal-2024-plausibly}, defined as \textit{\textbf{1-Impossible}, \textbf{2-Technically Possible}, \textbf{3-Plausible}, \textbf{4-Likely} and \textbf{5-Very Likely}.} 
Given a question $q$ and possible answer $a$, judges are asked to assess the plausibility of $a$ under each of the following four conditions:
\begin{itemize}[nosep]
    \item \textbf{\texttt{NO} Rationale}: The $(q, a)$ pair is presented alone. No rationale is provided.
    \item \textbf{\texttt{PRO} Rationale}: The $(q, a)$ pair is accompanied by an argument for the \textit{plausibility} of $a$, i.e., a \texttt{PRO} rationale.
    \item \textbf{\texttt{CON} Rationale}: The $(q, a)$ pair is accompanied by an argument for the \textit{implausibility} of $a$, i.e., a \texttt{CON} rationale.
    \item \textbf{\texttt{PRO+CON} Rationale}: The $(q, a)$ pair is accompanied by both a \texttt{PRO} rationale \textit{and} a \texttt{CON} rationale.
\end{itemize}

For the \texttt{No} Rationale setting, we use the plausibility judgments released by \citet{palta-etal-2024-plausibly}. For each of the three remaining rationale settings (\texttt{PRO}, \texttt{CON}, and \texttt{PRO+CON}) we collect $5$ judgments, totaling 15 new judgments per selected $(q,a)$ pair.

Annotators are recruited through Prolific, and compensated at $\$15$/hour. To ensure independence, each annotator could judge at most one $(q,a,r)$ pair for a given question $q$ for a rationale type $r$. Further details, including examples of the annotation interfaces (Figures~\ref{siqa_pro_interface},~\ref{cqa_con_interface} and~\ref{cqa_pro_con_interface}), are presented in Appendix~\ref{annotation_protocol}. We present the overall annotation counts in \autoref{tab:anno_stats}.

\subsection{How do Rationales influence Human Ratings?}\label{rationales_impact_humans}
We measure how mean plausibility ratings\footnote{Average rating across the 5 annotators for each $(q,a)$ item.} for $(q,a)$ pairs change between the original \texttt{NO} Rationale setting and each of the \texttt{PRO}, \texttt{CON}, and \texttt{PRO+CON} Rationale settings.
\autoref{tab:rating_changes_agents} shows that, for human judges, mean plausibility ratings often fall or stay unchanged---but rarely rise---in the presence of \texttt{CON} rationales. With \texttt{PRO} rationales, mean human ratings often stay unchanged, and are only slightly more likely to rise than drop. In the \texttt{PRO+CON} setting, ratings are unlikely to rise, but about equally likely to drop or stay unchanged.
\begin{table*}[t!]
\centering
\resizebox{\textwidth}{!}{
\begin{tabular}{|c|c|ccc|ccc|ccc|}
\toprule
\multirow{2}{*}{\textbf{Dataset}} & \multirow{2}{*}{\textbf{Agent}} 
& \multicolumn{3}{c|}{\textbf{Pro Rationale}} 
& \multicolumn{3}{c|}{\textbf{Con Rationale}} 
& \multicolumn{3}{c|}{\textbf{Pro+Con Rationales}} \\  
& & Overall & Gold Label & Distractor 
  & Overall & Gold Label & Distractor 
  & Overall & Gold Label & Distractor \\ 
\midrule
\multirow{3}{*}{SIQA} 
  & Human & \cellcolor{green!14} $3.33 (+0.1)$ & \cellcolor{red!14} $3.84 (-0.26)$ &  \cellcolor{green!14}$2.81 (+0.47)$ & \cellcolor{red!14} $2.28 (-0.94)$ & \cellcolor{red!14} $2.72 (-1.39)$ & \cellcolor{red!14} $ 1.85 (-0.49)$ & \cellcolor{red!14} $2.87 (-0.35)$ & \cellcolor{red!14} $3.36 (-0.75)$ & \cellcolor{green!14} $ 2.39 (+0.05)$ \\ \cmidrule{2-11}
  & OpenAI & \cellcolor{green!14} $3.7(+0.57)$ & \cellcolor{green!14} $4.25(+0.26)$ & \cellcolor{green!14} $3.14(+0.88)$ & \cellcolor{red!14} $1.9(-1.22)$ & \cellcolor{red!14} $2.28(-1.7)$ & \cellcolor{red!14} $1.52(-0.74)$ & \cellcolor{red!14} $2.9(-0.22)$ & \cellcolor{red!14} $3.38(-0.6)$ & \cellcolor{green!14} $2.42 (+0.16)$ \\ \cmidrule{2-11}
  & Non-OpenAI & \cellcolor{green!14} $3.6(+0.62)$ & \cellcolor{green!14} $3.94(+0.42)$ & \cellcolor{green!14} $3.26(+0.82)$ & \cellcolor{red!14} $2.21(-0.77)$ & \cellcolor{red!14} $2.42(-1.09)$ & \cellcolor{red!14} $1.99(-0.45)$ &  \cellcolor{red!14}$2.93(-0.05)$ & \cellcolor{red!14} $3.19(-0.32)$ & \cellcolor{green!14} $2.66(+0.22)$ \\ \midrule
\multirow{3}{*}{CQA}  
  & Human & $3.39 (0.0)$ & \cellcolor{red!14} $3.91 (-0.44)$ & \cellcolor{green!14} $2.86 (+0.45)$ & \cellcolor{red!14} $2.62 (-0.76)$ & \cellcolor{red!14} $ 3.28 (-1.08)$ & \cellcolor{red!14} $1.96 (-0.45)$ & \cellcolor{red!14} $2.97 (-0.41)$ & \cellcolor{red!14} $3.7 (-0.65)$ & \cellcolor{red!14} $2.24 (-0.17)$ \\ \cmidrule{2-11}
  & OpenAI & \cellcolor{green!14} $3.91(+0.64)$ & \cellcolor{green!14} $4.55 (+0.25)$ & \cellcolor{green!14} $3.27(+1.03)$ & \cellcolor{red!14} $2.02(-1.25)$ & \cellcolor{red!14} $2.56(-1.74)$ & \cellcolor{red!14} $1.48(-0.76)$ & \cellcolor{red!14} $3.12(-0.15)$ & \cellcolor{red!14} $3.76(-0.54)$ & \cellcolor{green!14} $2.49(+0.25)$ \\ \cmidrule{2-11}
  & Non-OpenAI & \cellcolor{green!14} $3.63(+0.61)$ & \cellcolor{green!14} $3.95(+0.31)$ & \cellcolor{green!14} $3.32(+0.92)$ & \cellcolor{red!14} $2.18(-0.84)$ & \cellcolor{red!14} $2.43(-1.21)$ & \cellcolor{red!14} $1.94(-0.47)$ & \cellcolor{red!14}$2.99(-0.03)$ & \cellcolor{red!14}$3.28(-0.36)$ & \cellcolor{green!14} $2.7(+0.3)$ \\ \bottomrule
\end{tabular}
}
\caption{Mean plausibility ratings given by Human, OpenAI, and Non-OpenAI agents after exposure to \texttt{PRO}, \texttt{CON}, and \texttt{PRO+CON} rationales, reported separately for $a_{gold-label}$ and $a_{distractor}$. Numbers outside parentheses indicate mean ratings after rationales; numbers in parentheses represent the corresponding changes ($\Delta$) from the No Rationale Plausibility Rating.}
\label{tab:mean_stats_agents}
\end{table*}
However, an interesting pattern emerges when we stratify outcomes based on whether the answer choice is a ``gold'' (correct) or ``distractor'' (incorrect) answer, as shown in \autoref{tab:mean_stats_agents}.
Critically, \textit{we observe that the effect of \textup{\texttt{PRO}} rationales on mean human judgments is bimodal}: on distractor answers, mean judgments rise, while for gold answers they fall. Thus, for implausible distractor answers, \texttt{PRO} rationales appear sufficiently persuasive to raise mean ratings. But why do ratings for gold answers drop? One possible explanation is rooted in Grice’s Maxim of Quantity \citep{grice1975logic}: a conversation should be as informative as possible, and not provide more information than what is required. When the \textsc{LLM} presents a rationale arguing that a highly obvious fact is ``plausible'', human judges might implicitly interpret the \texttt{PRO} rationales as arguing the answer is \textit{at most} plausible, rather than \textit{at least} plausible. This ``underselling'' introduces doubt, causing humans to actively downgrade their initial high-confidence ratings to align with the \textsc{LLM}-generated rationale.

With \texttt{CON} rationales, both gold and distractor answers show drops in mean plausibility, albeit with much larger drops for gold answers. (Interestingly, the drop for gold answers with \texttt{CON} rationales is over one full Likert scale point on average, the largest changes observed for any setting.) While boundary effects likely contribute to the differences observed between changes for gold and distractor labels (e.g., there is a cap on how much an already high rating can rise), they cannot account for differences between answers of the \textit{same} type (e.g. distractor) across different rationale settings. For example, we have seen that mean ratings for distractor answers rise with \texttt{PRO} rationales and fall with \texttt{CON} rationales; in our controlled experimental setup, the only non-random explanation for this difference is the rationale type itself. We present statistical significance tests at the end of this section, with a causal analysis of the factors that led to the changes in the ratings presented in \cref{sec:regression_analysis}.

For \texttt{PRO+CON} rationales, we observe an approximate averaging of the \texttt{PRO} and \texttt{CON} outcomes: for gold and distractor answers both, changes in mean ratings lie in between the changes observed in the \texttt{PRO} and \texttt{CON} settings, respectively.

We perform several Chi-squared tests of homogeneity to compare the rating distributions with and without rationales. For \textsc{SIQA}, p-values for the shifts in the distributions are $0.069$, $2.17E^{-9}$, and $0.001$ for the \texttt{PRO}, \texttt{CON}, and \texttt{PRO+CON} settings, respectively; for \textsc{CQA}, they are $0.001$, $9.24E^{-7}$, and $0.0006$. In nearly all cases, adding rationales yields statistically significant distribution shifts.

\begin{figure*}[!ht]
    \centering
    \includegraphics[width=\linewidth]{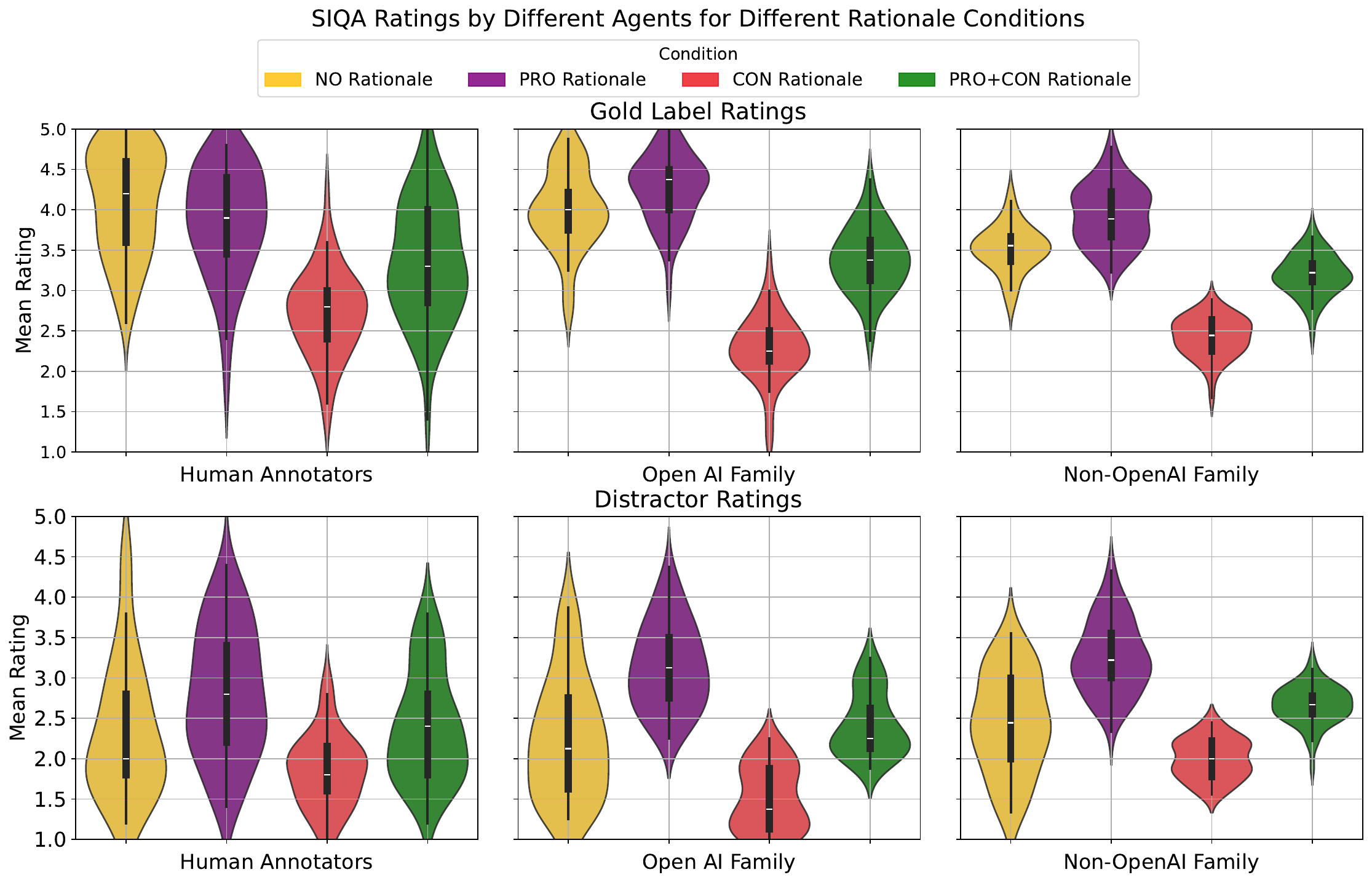}
    \caption{Mean Plausibility Rating for \textsc{SIQA} for Different Agents for Different Rationale Conditions.~\autoref{tab:mean_stats_agents} shows the mean changes for $a_{gold-label}$ and $a_{distractor}$ for all agents.}
    \label{fig:siqa_all_agent}
\end{figure*}
\section{Do Rationales Affect LLMs as well?}\label{sec:rationales_impact_llms}
In an effort to understand whether \textsc{LLM}s are also influenced by the introduction of different forms of rationales, we replicate the entire setup used to collect human plausibility ratings, as described in \cref{sec:human_ratings}.
Previous work \cite{panickssery2024llm} has shown that \textsc{LLM}s can recognize and favor their own generated outputs over outputs from other models. Given that \texttt{GPT-4o} generated the \texttt{PRO} and \texttt{CON} rationales, to mitigate any potential biases that can arise from instances of self-preference, we divide the \textsc{LLM}s we study in our work into two groups, with membership as follows:
\begin{itemize}[nosep]
    \item \textbf{OpenAI Models}: GPT (3.5, 4, 4-turbo, 4o, 4o-mini, 4.5-preview, o1, and o3-mini) \citep{openai2024gpt4technicalreport, openai2024gpt4ocard, openai2024openaio1card}
    \item \textbf{Non-OpenAI Models}: LLaMA-3.1-Instruct (8B, 70B) \citep{grattafiori2024llama3herdmodels}, LLaMa-3.2-Instruct (3B), LLaMa-3.3-Instruct (70B), Mistral Instruct (8x22B) \citep{jiang2024mixtralexperts}, Yi-1.5-Chat (6B, 9B, 34B) \citep{ai2025yiopenfoundationmodels} and DeepSeek-R1 (Qwen-32B) \citep{deepseekai2025deepseekr1incentivizingreasoningcapability}.
\end{itemize}
We prompt the models listed above with the $(q,a)$ pairs accompanied by one rationale for each setting $[\texttt{NO,PRO,CON,PRO+CON}]$ (same as used in human experiments), and present the zero-shot prompts in Prompts~\ref{prompt:without_rationale_rating}, \ref{prompt:pro_rationale_rating}, \ref{prompt:con_rationale_rating} and~\ref{prompt:pro_con_rationale_rating} in Appendix \ref{prompts}.

\subsection{LLM Ratings Pre and Post Rationales}
Paralleling our analysis in \S\ref{rationales_impact_humans}, we compare how mean plausibility ratings produced by LLMs  (OpenAI and Non-OpenAI models) change in the presence of each rationale type, reporting results again in Tables \ref{tab:rating_changes_agents} and \ref{tab:mean_stats_agents}, for direct comparison with results from the human judges.

Overall, \texttt{PRO} rationales have a larger positive impact for both sets of models, with a large proportion of ratings increasing across both datasets with the inclusion of the \texttt{PRO} rationale. Similarly, adding \texttt{CON} rationales leads to a large drop in ratings across both datasets. The inclusion of \texttt{PRO+CON} rationales once again elicits a mixed response, similar to our observation with human judgments in \cref{rationales_impact_humans}.

However, in sharp contrast to what we observed with human annotators, the introduction of \texttt{PRO} rationales leads to an increase in the mean plausibility ratings for $a_{gold-label}$ across both \textsc{SIQA} and \textsc{CQA} for both OpenAI and Non-OpenAI models. Similar to the shifts in human ratings for $a_{gold-label}$ and $a_{distractor}$, by adding \texttt{CON} rationales, both groups of models exhibit a decrease in mean plausibility ratings across both datasets.

\begin{figure*}[!ht]
    \centering
    \includegraphics[width=\linewidth]{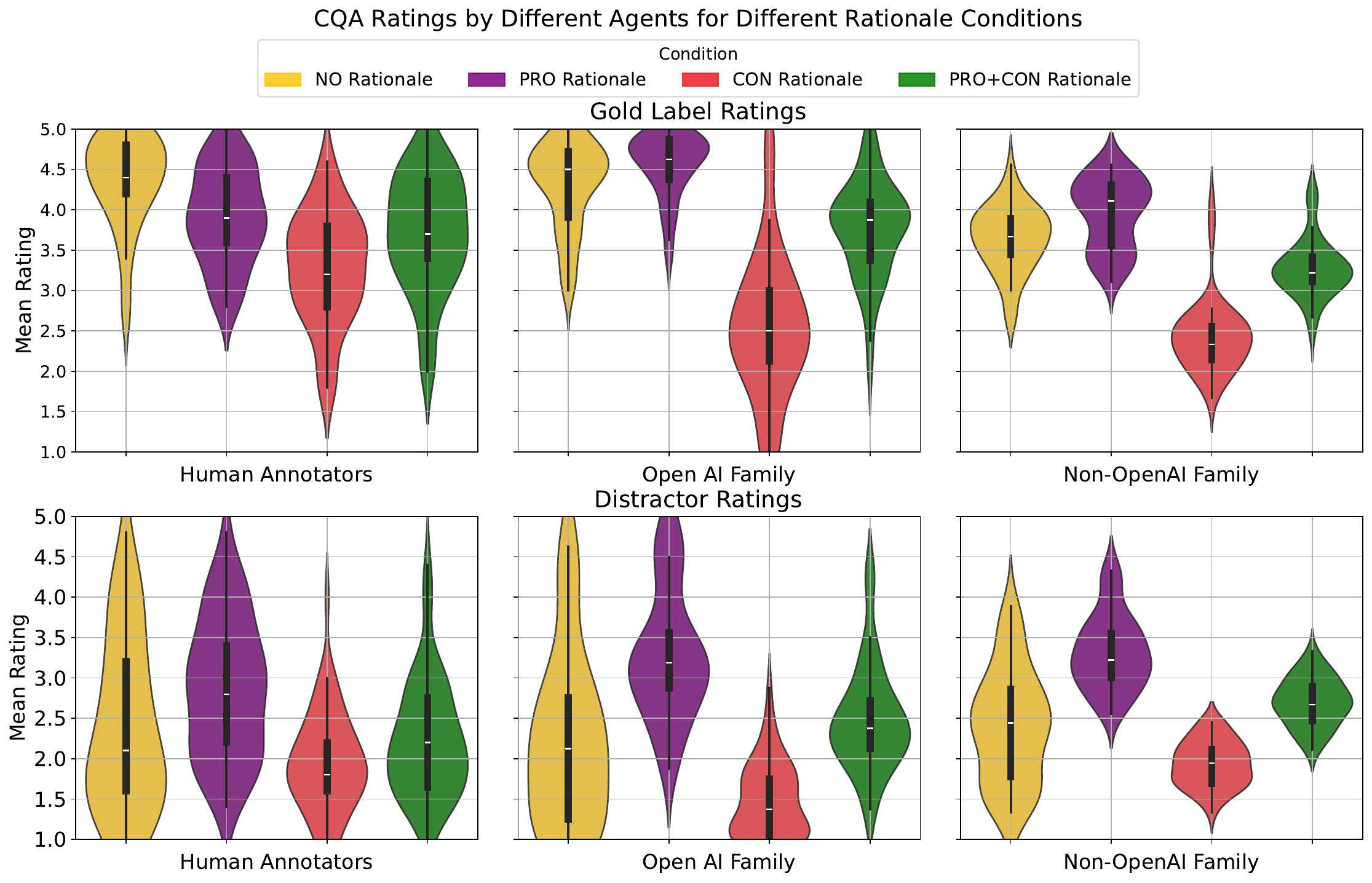}
    \caption{Mean Plausibility Rating for \textsc{CQA} for Different Agents for Different Rationale Conditions.~\autoref{tab:mean_stats_agents} shows the mean changes for $a_{gold-label}$ and $a_{distractor}$ for all agents.}
    \label{fig:cqa_all_agent}
\end{figure*}

Finally, when \texttt{PRO+CON} rationales are presented together with a $(q,a)$ item, we observe the \textsc{LLM}s to also display mixed results, with plausibility ratings for $a_{gold-label}$ dropping for both \textsc{SIQA} and \textsc{CQA}, but increasing for $a_{distractor}$. This behavior is similar to what was observed for the human ratings (\cref{rationales_impact_humans}). One possible explanation for this phenomenon is that, if \texttt{PRO} rationales dampen responses at the low end of the plausibility scale, and \texttt{CON} rationales dampen responses at the high end of the plausibility scale, then \texttt{PRO+CON} rationales together may have the effect of pushing plausibility ratings to the center of the scale because neither extreme is being used for rating the answer $a$. This would have the effect of raising average scores for low-scoring distractors $(a_{distractor})$ and lowering scores for high-scoring gold-label answers $(a_{gold-label})$, as is observed for both humans and \textsc{LLM}s. 
Figures~\ref{fig:siqa_all_agent} and~\ref{fig:cqa_all_agent} show that this is not simply a case of reversion to the mean, as, for instance, \texttt{CON} rationales on average push mean scores of distractors (which are low to begin with) even lower.
Another compatible explanation is that, as observed in human judgments, \texttt{PRO+CON} rationales result in shifts in mean judgments that lie between the shifts observed in the \texttt{PRO} and \texttt{CON} settings. 

To establish the significance of these shifts, we ran several Chi-squared tests of homogeneity to compare the ratings distributions with and without rationales within each setting. These shifts were found to be highly statistically significant across all settings, with all p-values $<0.0001$.

Comparing the ratings given by OpenAI and Non-OpenAI models, we see the OpenAI models to be more sensitive to the inclusion of \texttt{PRO}, \texttt{CON} and \texttt{PRO+CON} rationales, which we hypothesize to be because of the OpenAI models favoring the rationales generated by \texttt{GPT-4o}. 

We include examples from \textsc{SIQA} and \textsc{CQA} for each of \texttt{PRO} rationales (\autoref{pro_examples}), \texttt{CON} rationales (\autoref{con_examples}) and \texttt{PRO+CON} rationales (Tables~\ref{pro_con_examples} and~\ref{pro_con_examples_contd}), and show how human and \textsc{LLM} ratings differ when different rationales are included for a $(q,a)$ item as compared to annotating without any rationales. 
\section{Why Did the Ratings Change?}\label{sec:regression_analysis}
We have now established that both humans and \textsc{LLM}s are sensitive to the inclusion of \texttt{PRO}, \texttt{CON} and \texttt{PRO+CON} rationales, and adjust their plausibility ratings accordingly. To better understand what factors contributed the most to the changes in the plausibility ratings, we perform ordinary least squares (\textsc{OLS}) regression on the collected data separately for $a_{gold-label}$ and $a_{distractor}$ using the rationale type (\texttt{PRO,CON,PRO+CON}) and the \texttt{NO} Rationale Plausibility Rating as features, with the dependent variable being the change $(\Delta)$ in plausibility rating after a rationale is presented. The \texttt{NO} Rationale Plausibility Rating was originally collected on the 1-5 Likert Scale as described in \citet{palta-etal-2024-plausibly}. For the regression, we use a $[-2,2]$ scale to center the ratings around $0$.

As detailed in \autoref{tab:regression}, the \texttt{NO} Rationale Plausibility Rating is a statistically significant predictor ($p<0.05$) of the rating change for humans, OpenAI and Non-OpenAI models. The consistently negative coefficients reveal a strong anchoring effect: the higher the initial plausibility of an answer, the smaller the subsequent change in its rating after a rationale is provided. This suggests that all rater groups are less likely to raise their opinion if their initial assessment is already high. Additionally, this anchoring effect is more pronounced for distractors than for gold-label answers across all groups. This indicates that the initial rating of an incorrect answer choice has a much stronger influence on the final rating change than the initial rating about the correct one. 

Comparing the different rationale types included with the $(q,a)$ items, we find that \texttt{CON} rationales exert a stronger effect than \texttt{PRO} rationales, with large negative coefficients that sharply \textit{reduce} plausibility for otherwise correct answers. \texttt{PRO} rationales do \textit{increase} plausibility, but the magnitude of their coefficients is smaller, indicating that supporting reasons reinforce less than opposing reasons undermine. \texttt{PRO+CON} rationales typically carry small negative coefficients, indicating that when both supporting and opposing arguments are presented together, the opposing component tends to dominate, yielding a net plausibility decrease.
\begin{table*}[htbp!]
\centering
\small
\resizebox{\textwidth}{!}{
\begin{tabular}{|cc|cccccccc|}
\cline{3-10}
\multicolumn{1}{l}{}& \multicolumn{1}{l|}{} & \multicolumn{8}{c|}{\textbf{Feature}} \\
\multicolumn{1}{l}{}&& \multicolumn{2}{c|}{\textbf{\texttt{NO} Rationale Rating}}& \multicolumn{2}{c|}{\textbf{\texttt{PRO} Rationale}}& \multicolumn{2}{c|}{\textbf{\texttt{CON} Rationale}}& \multicolumn{2}{c|}{\textbf{\texttt{PRO+CON} Rationale}} \\ \midrule
\multicolumn{1}{|c|}{\textbf{Dataset}}& \textbf{Agent}& Gold Label& \multicolumn{1}{c|}{Distractor} & Gold Label & \multicolumn{1}{c|}{Distractor} & Gold Label & \multicolumn{1}{c|}{Distractor} & Gold Label & Distractor\\ \midrule
\multicolumn{1}{|c|}{\multirowcell{3}{SIQA}} & Human & $-\textbf{0.4581}$ & \multicolumn{1}{c|}{$-\textbf{0.5495}$} & $\textbf{0.2417}$ & \multicolumn{1}{c|}{$0.1082$} & $-\textbf{0.8823}$ & \multicolumn{1}{c|}{$-\textbf{0.8525}$} & $-\textbf{0.2423}$ & $-\textbf{0.3125}$ \\ \cmidrule{2-10}
\multicolumn{1}{|c|}{} & OpenAI & $-\textbf{0.4503}$ & \multicolumn{1}{c|}{$-\textbf{0.5086}$} & $\textbf{0.7083}$ & \multicolumn{1}{c|}{$\textbf{0.5023}$} & $-\textbf{1.2592}$ & \multicolumn{1}{c|}{$-\textbf{1.1177}$} & $-\textbf{0.1592} $ & $-\textbf{0.2127}$ \\ \cmidrule{2-10}
\multicolumn{1}{|c|}{} & Non-OpenAI & $-\textbf{0.4770}$ & \multicolumn{1}{c|}{$-\textbf{0.6491}$} & $\textbf{0.6659}$ & \multicolumn{1}{c|}{$\textbf{0.4543}$} & $-\textbf{0.8452}$ & \multicolumn{1}{c|}{$-\textbf{0.8102}$} & $-0.0785$ & $-\textbf{0.1391}$ \\ \midrule
\multicolumn{1}{|c|}{\multirow{3}{*}{CQA}} & Human & $-\textbf{0.5092}$ & \multicolumn{1}{c|}{$-\textbf{0.4983}$} & $0.2445$ & \multicolumn{1}{c|}{$\textbf{0.1590}$} & $-\textbf{0.3875}$ & \multicolumn{1}{c|}{$-\textbf{0.7450}$} & $0.0405$ & $-\textbf{0.4650}$ \\ \cmidrule{2-10}
\multicolumn{1}{|c|}{} & OpenAI & \multicolumn{1}{c}{$-\textbf{0.2340}$} & \multicolumn{1}{c|}{$-\textbf{0.4883}$} & $\textbf{0.5517}$ & \multicolumn{1}{c|}{$\textbf{0.6614}$} & $-\textbf{1.4408}$ & \multicolumn{1}{c|}{$-\textbf{1.1286}$} & $-0.2383$ & $-\textbf{0.1236}$ \\ \cmidrule{2-10} 
\multicolumn{1}{|c|}{} & Non-OpenAI & \multicolumn{1}{c}{$-\textbf{0.5632}$} & \multicolumn{1}{c|}{$-\textbf{0.6236}$} & $\textbf{0.6693}$ & \multicolumn{1}{c|}{$\textbf{0.5428}$} & $-\textbf{0.8507}$ & \multicolumn{1}{c|}{$-\textbf{0.8394}$} & $-0.0040$ & $-0.0750$ \\ \bottomrule
\end{tabular}
}
\caption{OLS Regression Coefficients for the \texttt{NO} Rationale Plausibility Rating and Rationale Type (\texttt{PRO}, \texttt{CON} and \texttt{PRO+CON}) features. The dependent variable here is the change $(\Delta)$ in the plausibility rating after a rationale was presented. Numbers highlighted in \textbf{bold} represent coefficients that are statistically significant $(p<0.05)$ according to the t-test.}
\label{tab:regression}
\end{table*}

Finally, we also test the rationale length, measured as the number of words in the generated rationale, as a feature in a separate regression where the dependent variable is the magnitude of the change in the plausibility rating, and present the results in \autoref{tab:pre_rationale}. Across all settings, the regression coefficients are close to zero suggesting that rationale length exhibits only a weak linear relationship with rating scores, regardless of the answer choice type, dataset or the annotating agent.
\section{Related Works}\label{related}
Rating the likelihood or plausibility of events in language has been explored previously in commonsense reasoning \citep{wang-etal-2018-modeling, liu-etal-2023-vera}. The graded nature \citep{zhang-etal-2017-ordinal, chen-etal-2020-uncertain}, the subjectivity of commonsense reasoning \citep{doi:10.1073/pnas.2309535121} along axes like culture \citep{palta-rudinger-2023-fork, hershcovich-etal-2022-challenges, bhatia-shwartz-2023-gd, acquaye-etal-2024-susu} has been long studied. We take advantage of this subjectivity to introduce \textsc{LLM}-generated rationales into a commonsense \textsc{MCQ} item and see the impact on human and \textsc{LLM} ratings. Previous works have also looked at how biases in human probability estimates \citep{bias-heuristics, kruglanski1983bias, Martins_2006}, framing effects \citep{DRUCKMAN200191, mandel2014framing, sher2008framing}, recency bias \citep{jones1976recency, steiner1989immediate, arnold2000effect} and availability heuristics \citep{TVERSKY1973207, availability_prof, Availability_risk, pachur2012people, availability_consumer} can impact human decision making.

\textsc{LLM}s can be trained or prompted to produce rationales of many forms \citep{wiegreffe2021teach}; early work on training \textsc{LMs} to produce free-form textual rationales focused on explaining model predictions on computer vision or natural language inference tasks \citep{hendricks2016generating,NEURIPS2018_4c7a167b}. \textsc{LMs}' ability to generate free-form textual explanations for commonsense reasoning tasks was first established by \citet{rajani-etal-2019-explain, shwartz-etal-2020-unsupervised, latcinnik2020explaining, Brahman_Shwartz_Rudinger_Choi_2021, wiegreffe-etal-2021-measuring}, with subsequent work showing that large pre-trained \textsc{LMs} could generate high-quality explanations for commonsense tasks without having been explicitly trained to do so \citep{paranjape-etal-2021-prompting, lampinen-etal-2022-language, wiegreffe-etal-2022-reframing, marasovic-etal-2022-shot}.

It is now widely accepted that \textsc{LMs} can produce fluent textual explanations for a wide variety of NLP tasks. The role of these explanations on human (over-)trust, reliance \citep{10.1145/3442188.3445923, goyal-etal-2023-else}, distorting beliefs \citep{batista2026rationalanalysiseffectssycophantic}, and utility to humans doing tasks has been studied in a variety of task settings \citep{10.5555/3172077.3172090, 10.1145/3301275.3302265, gonzalez-etal-2021-explanations, 10.1145/3411764.3445717, parrish-etal-2022-single, lee2023evaluating, 10.1145/3593013.3593970, joshi-etal-2023-machine, si-etal-2024-large, fok2024search}, though without a specific focus on understanding how explanations affect humans' perception of the plausibility of commonsense situations (which can be inherently ambiguous), a domain where humans are traditionally considered to be the ``experts''. 

To the best of our knowledge, our finding about the bimodal effect of \texttt{PRO} rationales on human judges: raising the ratings for implausible distractor answers but dropping the ratings for highly plausible gold-label answers has not been documented in prior works on \textsc{LLM}-generated explanations. Further, our causal \textsc{OLS} regression analysis also confirms a strong anchoring effect (a higher initial plausibility leads to a smaller change in the ratings) for all rater groups (humans and \textsc{LLM}s).The finding that this anchoring effect is more pronounced for distractors than for gold-label answers for all groups is a new insight into the rigidity of initial judgments for incorrect answers.
\section{Conclusion}
In this work, we have tried to answer the critical question, \textit{How do explanations work differently depending on the context, the audience (human vs. LLMs), and the baseline certainty of the claim?} By collecting $3000$ human and $13,600$ \textsc{LLM} ratings for $200$ $(q,a)$ pairs, we are able to demonstrate that both human and \textsc{LLM} plausibility ratings shift when rationales either in favor, against, or presented together are included with a commonsense reasoning question $q$ and answer $a$, highlighting the potentially persuasive nature of \textsc{LLM}-generated explanations, specifically for commonsense reasoning tasks. Our findings also highlight that humans and \textsc{LLM}s do not always react similarly when shown different rationales, with models showing a higher sensitivity to the rationales, underscoring the challenges of aligning AI-generated judgments with human intuition, particularly in subjective reasoning tasks. Finally, our findings highlight the potential for \textsc{LLM}s to aid humans by challenging users' reasoning, stress-testing arguments, and introducing alternative perspectives, all of which have direct implications on effective human-\textsc{AI} collaboration, human-\textsc{AI} safety, \textsc{AI} for decision support, and the methodology of how commonsense reasoning is evaluated in \textsc{AI} systems. However, the fact that \textsc{LLM}-generated rationales are able to exert considerable influence on human annotators' beliefs, even in a domain like commonsense reasoning where humans are traditionally considered to be the ``experts'', also raises the question that if \textsc{LLM}s can sway human judgments on basic intuitive facts, the risk of persuasive manipulation or psychological influence might be higher in specialized or technical domains. We also demonstrate that \textsc{LLM}s can effectively steer human judgment toward more implausible inferences through persuasive argumentation. This highlights a critical vulnerability in human-\textsc{AI} collaboration: users may prioritize the plausibility of an \textsc{LLM}'s argument over their own knowledge and beliefs. Because it is somewhat counter-intuitive that \textsc{LLM} reasoning can influence human reasoning in the domain of everyday commonsense reasoning, users may also be less likely to realize when such manipulations have occurred. Addressing these risks requires robust safeguards, including transparency, bias mitigation, and mechanisms to detect and counteract harmful manipulative uses. Future works will focus on how factors such as cognitive abilities, age, grade-level and other demographic characteristics can influence human responses to \textsc{LLM}-generated explanations.
\section*{Limitations}
Our goal in this work was to determine whether \textit{any} \textsc{LLM} might produce rationales capable of swaying human (and other LLMs') plausibility judgments. For that purpose, \texttt{GPT-4o} proved effective at creating convincing arguments which, as we demonstrated, led to the changes in plausibility ratings. However, the persuasive power and style of \textsc{LLM}-generated rationales can vary significantly across different models and architectures. We hope that our findings inspire future works to look into the extent to which different models influence human and \textsc{LLM} plausibility judgments. 

The scope of our investigation is confined to two English-language commonsense reasoning datasets, Social IQA and CommonsenseQA. The plausibility of an answer choice and the extent to which a rationale is persuasive might differ in other languages and domains, such as technical, scientific, or legal reasoning, where arguments may be more constrained by factual evidence. Future work may explore understanding the impact of \textsc{LLM}-generated rationales on domains beyond commonsense reasoning. Furthermore, our human annotators were recruited from a specific demographic (US-based, English-speaking users on Prolific). As common sense itself can be culturally situated, the impact of these rationales could vary across different cultural and linguistic groups. While we employed rigorous annotation protocols and made sure each instance was annotated by at least $5$ annotators, there may still be instances where ambiguity in interpretation arises and hence, the impact a rationale has on the plausibility ratings would be different.
\section{Acknowledgements}
We would like to thank the anonymous reviewers for their valuable feedback on this paper. We would also like to thank Hal Daumé III, Marine Carpuat, Naomi Feldman, Srajit Sakhuja, Abhilasha Sancheti, Haozhe An, Neha Srikanth, Christabel Acquaye, Yu Hou, Nishant Balepur and other members of the \abr{CLIP} Lab for their helpful comments and suggestions. Rachel Rudinger and Shramay Palta are supported by \abr{NSF CAREER} Award No.~2339746. Any opinions, findings, and conclusions or recommendations expressed in this material are those of the author(s) and do not necessarily reflect the views of the National Science Foundation.
\bibliography{bib/custom}
\appendix
\section{Appendix}\label{sec:appendix}
\begin{table*}[t]
\centering
\small
\begin{tabular}{|c|cc|cc|cc|}
\toprule
\multirow{2}{*}{\textbf{Dataset}} & \multicolumn{2}{c|}{\textbf{Humans}} & \multicolumn{2}{c|}{\textbf{OpenAI Models}} & \multicolumn{2}{c|}{\textbf{Non-OpenAI Models}} \\  
& Gold Label & Distractor & Gold Label & Distractor & Gold Label & Distractor \\ \midrule
SIQA & $-0.0025$ & $-0.0021$ & $-\textbf{0.0068}$ & $-\textbf{0.0058}$ & $-\textbf{0.0074}$ & $-0.0023$ \\ \midrule
CQA & $-0.0042$ & $-\textbf{0.0037}$ & $-\textbf{0.0113}$ & $-\textbf{0.0046}$ & $-\textbf{0.0081}$ & $-\textbf{0.0037}$ \\ \bottomrule
\end{tabular}
\caption{OLS regression coefficients for the Rationale Length as a feature for ratings given by Humans, OpenAI and Non-OpenAI models. The dependent variable here is the change $(\Delta)$ in the plausibility rating after a rationale was presented. Numbers highlighted in \textbf{bold} represent coefficients that are statistically significant $(p<0.05)$ according to the t-test.}
\label{tab:pre_rationale}
\end{table*}
\subsection{License for Artifacts}
All datasets used in this work are publicly available and free to use on HuggingFace. All \textsc{LLM}s are used in accordance with the licenses released by their respective developing organizations.

\subsection{Usage of AI Assistants}
We only use AI Assistants to assist our writing to identify grammar errors, typos and rephrase terms for readability.

\subsection{Computational Details}
LLaMA-3.1-Instruct (8B, 70B), LLaMA-3.2-Instruct (3B), LLaMA-3.3-Instruct (70B), Mistral Instruct (8x22B), Yi-1.5-Chat (6B, 9B, 34B) and DeepSeek-R1 (Qwen 32B) were all run on eight NVIDIA:RTXA5000 GPUs and were allocated a total of $24$ GPU hours to run all experiments. All OpenAI models were run on CPU and were allocated a total of $8$ hours to run all experiments. Each \textsc{LLM} decodes with a maximum token generation length of 50, to ensure no extra verbiage is generated. We did not perform a hyperparameter search. All results are obtained from a single run. 

For generating the rationales as mentioned in \cref{rationale_generation}, we set temperature to be $1$ for GPT-4o. For generating plausibility ratings as discussed in \cref{sec:rationales_impact_llms}, we set temperature to be $0$ and $0.1$ for OpenAI models and Non-OpenAI models respectively to ensure greedy decoding.
\subsection{Annotation Process Details}\label{annotation_protocol}
We used Prolific to collect the human annotations for all three rationale conditions. The annotators for our task were selected on the basis of the following criteria:
\begin{enumerate}[nosep]
    \item Must be located in the United States.
    \item Primary language must be English.
    \item Must not have any literacy difficulties.
    \item Must have attained a minimum of an undergraduate level degree.
    \item Must have an approval rate between $99-100\%$ on Prolific.
    \item We use a $50-50$ split of male and female\footnote{Gender as indicated on Prolific.} annotators to minimize the risk of any gender-specific biases creeping in.
\end{enumerate}
The total cost for our entire human annotations for all three rationales came out to be $\$1000$. We also received an exempt status from the \textsc{IRB} at our institution for this research. We do not collect any personally identifiable information from the annotators.

\subsection{Prompts used for Rationale Generation and Plausibility Ratings}\label{prompts}
We release the prompts used in our work for generating the rationales  and prompting the models for plausibility ratings below:
\begin{prompt}[title={Prompt \thetcbcounter: PRO Rationale Generation Prompt}, label=prompt:rationale_generation_pro]
\texttt{Give your best rationale as to why the given choice is a plausible answer for the question. The rationale should not be more than 2 sentences long and should be concise. Your rationale has to be in favour of the plausibility of the choice, no matter what. Some of these situations may require creativity or thinking outside to box to come up with a rationale that best explains the plausibility of the answer choice. Feel free to be as creative as required.}\\
\texttt{Question:} $q$\\
\texttt{Choice:} $a$\\
\texttt{Generated Rationale:} \colorbox{\colorBoxColorPro}{\textcolor{white}{\texttt{PRO} Rationale}}
\end{prompt}

\begin{prompt}[title={Prompt \thetcbcounter: CON Rationale Generation Prompt}, label=prompt:rationale_generation_con]
\texttt{Give your best rationale as to why the given choice is an implausible answer for the question. The rationale should not be more than 2 sentences long and should be concise. Your rationale has to be in favour of the implausibility of the choice, no matter what. Some of these situations may require creativity or thinking outside to box to come up with a rationale that best explains the implausibility of the answer choice. Feel free to be as creative as required.}\\
\texttt{Question:} $q$\\
\texttt{Choice:} $a$\\
\texttt{Generated Rationale:} \colorbox{\colorBoxColorCon}{\textcolor{white}{\texttt{CON} Rationale}}
\end{prompt}

\begin{prompt}[title={Prompt \thetcbcounter: Plausibility Rating Without Rationales}, label=prompt:without_rationale_rating]
\texttt{I will give you a question and an answer choice. Your task is to rate the plausibility of the answer choice on the following Likert scale: \\
Likert Scale:\\
1 - Impossible\\
2 - Technically Possible\\
3 - Plausible\\
4 - Likely\\
5 - Very Likely\\
Your answer should just be a complete option from the Likert scale. Nothing else. No extra words or explanations.}\\
\texttt{Question:} $q$\\
\texttt{Choice:} $a$\\
\texttt{Generated Rating:} \colorbox{\colorBoxColor}{$r$}
\end{prompt}

\begin{prompt}[title={Prompt \thetcbcounter: Plausibility Rating With Pro Rationale}, label=prompt:pro_rationale_rating]
\texttt{I will give you a question, an answer choice, and a rationale that best explains the plausibility of the answer choice. Your task is to rate the plausibility of the answer choice on the following Likert scale taking into consideration the rationale given. \\
Likert Scale:\\
1 - Impossible\\
2 - Technically Possible\\
3 - Plausible\\
4 - Likely\\
5 - Very Likely\\
Your answer should just be a complete option from the Likert scale. Nothing else. No extra words or explanations.}\\
\texttt{Question:} $q$\\
\texttt{Choice:} $a$\\
\texttt{Explanation}: \textcolor{purple_plot}{\texttt{PRO} Rationale}\\
\texttt{Generated Rating:} \colorbox{\colorBoxColorPro}{\textcolor{white}{$r$}}
\end{prompt}

\begin{prompt}[title={Prompt \thetcbcounter: Plausibility Rating With Con Rationale}, label=prompt:con_rationale_rating]
\texttt{I will give you a question, an answer choice, and a rationale that best explains the implausibility of the answer choice. Your task is to rate the plausibility of the answer choice on the following Likert scale taking into consideration the rationale given. \\
Likert Scale:\\
1 - Impossible\\
2 - Technically Possible\\
3 - Plausible\\
4 - Likely\\
5 - Very Likely\\
Your answer should just be a complete option from the Likert scale. Nothing else. No extra words or explanations.}\\
\texttt{Question:} $q$\\
\texttt{Choice:} $a$\\
\texttt{Explanation}: \textcolor{UMDred}{\texttt{Con} Rationale}\\
\texttt{Generated Rating:} \colorbox{\colorBoxColorCon}{\textcolor{white}{$r$}}
\end{prompt}

\begin{prompt}[title={Prompt \thetcbcounter: Plausibility Rating With Pro+Con Rationales}, label=prompt:pro_con_rationale_rating]
\texttt{I will give you a question, an answer choice, a rationale that best explains the plausibility of the answer choice, and a rationale that best explains the implausibility of the answer choice. Your task is to rate the plausibility of the answer choice on the following Likert scale taking into consideration the rationales given. \\
Likert Scale:\\
1 - Impossible\\
2 - Technically Possible\\
3 - Plausible\\
4 - Likely\\
5 - Very Likely\\
Your answer should just be a complete option from the Likert scale. Nothing else. No extra words or explanations.}\\
\texttt{Question:} $q$\\
\texttt{Choice:} $a$\\
\texttt{Explanation 1}: \textcolor{purple_plot}{\texttt{Pro} Rationale}\\
\texttt{Explanation 2}: \textcolor{UMDred}{\texttt{Con} Rationale}\\
\texttt{Generated Rating:} \colorbox{\colorBoxColorProCon}{\textcolor{white}{$r$}}
\end{prompt}

\clearpage

\begin{table*}[ht]
\centering
\begin{tabularx}{\linewidth}{|c | X |}
\toprule
\textbf{Agent Rating Change} & \multicolumn{1}{c|}{\textbf{Example}} \\ \midrule
\specialcell{Human Rating $\uparrow$ \\ \\ LLM Rating $\uparrow$}   & \specialcellleft{\textit{Context}: Jenny wanted to found out what happened to Quinn's \\leg and read Quinn's book about his travels where he \\discussed his accident later in the first chapter.\\
\textit{Question}: How would Jenny feel as a result? \\
\textit{Choice}: satisfied \textit{(gold label)}  \\ \textit{Pro Rationale:} Jenny would feel satisfied because she \\successfully found the information she was seeking about \\Quinn's leg accident in his book, fulfilling her curiosity and \\providing her with the answers she needed. \\ 
Human Rating $\Delta$: $+1.2$ \\
LLM Rating $\Delta$ (OpenAI): $+0.88$ \\
LLM Rating $\Delta$ (Non-OpenAI): $+0.67$
}\\ \midrule
\specialcell{Human Rating $\uparrow$ \\ \\ LLM Rating $\downarrow$} & \specialcellleft{\textit{Question}: I took my seat, the curtains drew back and I enjoyed \\the what? \\
\textit{Choice}: movie \textit{(distractor)}  \\ \textit{Pro Rationale:} The mention of "curtains drew back" is \\commonly associated with the start of a film in a theater \\setting, making "movie" a plausible choice as to what was \\enjoyed. Additionally, sitting down right before an experience \\occurs supports the idea of a movie theater viewing.\\ 
Human Rating $\Delta$: $+0.2$ \\
LLM Rating $\Delta$ (OpenAI): $-0.12$ \\
LLM Rating $\Delta$ (Non-OpenAI): $-0.22$
} \\ \midrule

\specialcell{Human Rating $\downarrow$ \\ \\ LLM Rating $\uparrow$} & \specialcellleft{\textit{Context}: Quinn got a new job in a different town so they \\moved to a new house.\\
\textit{Question}: What will Quinn want to do next? \\
\textit{Choice}: reload his things \textit{(distractor)}  \\ \textit{Pro Rationale:} Quinn may want to reload his things as part of \\the process of unpacking and arranging his belongings in his \\new home, ensuring everything is set up correctly and \\efficiently after the move. Additionally, reloading could refer \\to reorganizing or reconfiguring items like furniture and \\electronics to better suit the new living space. \\ 
Human Rating $\Delta$: $-1.2$ \\
LLM Rating $\Delta$ (OpenAI): $+1$ \\
LLM Rating $\Delta$ (Non-OpenAI): $+1$
} \\
\midrule

\specialcell{Human Rating $\downarrow$ \\ \\ LLM Rating $\downarrow$} & \specialcellleft{\textit{Question}: The teacher played on the upright piano, she was \\explaining the song to all the students in the what? \\
\textit{Choice}: music room \textit{(gold label)}  \\ \textit{Pro Rationale:} The choice "music room" is plausible because it's \\a common setting within a school where pianos are typically \\located and where a teacher would explain music to students, \\making it a logical location for the scenario. \\ 
Human Rating $\Delta$: $-1$ \\
LLM Rating $\Delta$ (OpenAI): $-0.62$ \\
LLM Rating $\Delta$ (Non-OpenAI): $-0.33$
} \\
\bottomrule
\end{tabularx}
\caption{Examples from \textsc{SIQA} and \textsc{CQA} showing mean plausibility rating changes for different agents for \texttt{PRO Rationales.}}\label{pro_examples}
\end{table*}

\begin{table*}[ht]
\centering
\begin{tabularx}{\linewidth}{|c | X |}
\toprule
\textbf{Agent Rating Change} & \multicolumn{1}{c|}{\textbf{Example}} \\ \midrule
\specialcell{Human Rating $\uparrow$ \\ \\ LLM Rating $\uparrow$}   & \specialcellleft{\textit{Question}: Where can many stores with clothing be found? \\
\textit{Choice}: library \textit{(distractor)}  \\ \textit{Con Rationale:} While a mall is a common place for multiple \\clothing stores, it is implausible to consider it the specific \\answer because a fashion district or online marketplaces might \\have far more clothing stores concentrated in one place. \\ 
Human Rating $\Delta$: $+0.6$ \\
LLM Rating $\Delta$ (OpenAI): $0$ \\
LLM Rating $\Delta$ (Non-OpenAI): $+0.33$
}\\ \midrule
\specialcell{Human Rating $\uparrow$ \\ \\ LLM Rating $\downarrow$} & \specialcellleft{\textit{Context}: Cameron wasted money on an expensive handbag.\\
\textit{Question}: Why did Cameron do this? \\
\textit{Choice}: resent her waste \textit{(distractor)}  \\ \textit{Con Rationale:} It is implausible that Cameron would waste \\money on an expensive handbag with the intention of \\resenting her waste because typically, people do not make \\financial decisions to purposefully feel negative emotions like \\resentment. Such a motive is counterproductive and irrational, \\as purchases are usually made for positive reasons such as \\necessity, desire, or status, not to induce regret.\\ 
Human Rating $\Delta$: $+1.6$ \\
LLM Rating $\Delta$ (OpenAI): $-0.62$ \\
LLM Rating $\Delta$ (Non-OpenAI): $-0.11$
} \\ \midrule

\specialcell{Human Rating $\downarrow$ \\ \\ LLM Rating $\downarrow$} & \specialcellleft{\textit{Context}: Riley layered down their arms with a blanket to keep \\warm.\\
\textit{Question}: What does Riley need to do before this? \\
\textit{Choice}: get a blanket from the closet \textit{(gold label)}  \\ \textit{Con Rationale:} It is implausible that Riley needs to get a blanket \\from the closet because the sentence implies that the blanket is \\already in their possession, suggesting that any necessary \\retrieval must have already occurred. Additionally, there is no \\indication that Riley is not near a blanket or lacks access to one \\already placed conveniently nearby. \\ 
Human Rating $\Delta$: $-3.2$ \\
LLM Rating $\Delta$ (OpenAI): $-1.38$ \\
LLM Rating $\Delta$ (Non-OpenAI): $-0.78$
} \\
\bottomrule
\end{tabularx}
\caption{Examples from \textsc{SIQA} and \textsc{CQA} showing mean plausibility rating changes for different agents for \texttt{CON Rationales.} We observed no cases where there was a decrease in Human ratings, but decrease in LLM ratings.}\label{con_examples}
\end{table*}

\begin{table*}[ht]
\centering
\begin{tabularx}{\linewidth}{|c | X |}
\toprule
\textbf{Agent Rating Change} & \multicolumn{1}{c|}{\textbf{Example}} \\ \midrule
\specialcell{Human Rating $\uparrow$ \\ \\ LLM Rating $\uparrow$}   & \specialcellleft{\textit{Context}: Aubrey was a police officer that pulled a driver over \\for speeding down the highway one day.\\
\textit{Question}: What will Aubrey want to do next? \\
\textit{Choice}: run away from the driver \textit{(distractor)}  \\ \textit{Pro Rationale:} Aubrey might want to run away from the driver \\if the situation escalates into a dangerous confrontation or if \\they perceive an immediate threat to their safety, prompting \\them to seek cover or help. As a police officer, ensuring \\personal safety can sometimes take precedence in volatile \\situations. \\ 
\textit{Con Rationale:} Running away from the driver is implausible \\because as a police officer, Aubrey's duty is to address the \\traffic violation and ensure public safety, not to avoid the \\situation. Such an action would be unprofessional and contrary \\to law enforcement responsibilities. \\ 
Human Rating $\Delta$: $+0.8$ \\
LLM Rating $\Delta$ (OpenAI): $+0.88$ \\
LLM Rating $\Delta$ (Non-OpenAI): $+1.11$
}\\ \midrule
\specialcell{Human Rating $\uparrow$ \\ \\ LLM Rating $\downarrow$} & \specialcellleft{\textit{Context}: Kendall took Skylar into acount when making plans \\for the day.\\
\textit{Question}: How would Skylar feel as a result? \\
\textit{Choice}: Glad they were thought about \textit{(gold label)}  \\ \textit{Pro Rationale:} Skylar would likely feel glad they were thought\\ about because being considered in someone's plans can lead to \\feelings of being valued and appreciated, reinforcing a sense of \\importance in their relationship with Kendall. This \\consideration shows that Kendall cares about Skylar's \\preferences and needs, which can foster a positive emotional \\response.\\
\textit{Con Rationale:} The choice is implausible because Skylar might\\ have preferred to be surprised rather than having their \\preferences considered, as they enjoy spontaneity and \\unpredictability. Additionally, Skylar may have wanted a day \\without any plans, making the consideration unnecessary and \\possibly unwanted.\\
Human Rating $\Delta$: $+0.2$ \\
LLM Rating $\Delta$ (OpenAI): $-1.12$ \\
LLM Rating $\Delta$ (Non-OpenAI): $-0.33$
} \\ 
\bottomrule
\end{tabularx}
\caption{Examples from \textsc{SIQA} and \textsc{CQA} showing mean plausibility rating changes for different agents when \texttt{PRO Rationales} and \texttt{CON Rationales} are shown together.}\label{pro_con_examples}
\end{table*}

\begin{table*}[ht]
\centering
\begin{tabularx}{\linewidth}{|c | X |}
\toprule
\textbf{Agent Rating Change} & \multicolumn{1}{c|}{\textbf{Example}} \\ \midrule
\specialcell{Human Rating $\downarrow$ \\ \\ LLM Rating $\uparrow$} & \specialcellleft{
\textit{Question}: James loved to surf but he wasn't good at it. He \\would always do what?\\
\textit{Choice}: enjoy yourself \textit{(distractor)}  \\ \textit{Pro Rationale:} Even though James wasn't skilled at surfing, his \\passion and love for it ensured that he focused more on the \\experience and thrill of being in the water, allowing him to \\enjoy himself regardless of his performance. This positive \\attitude highlights that enjoyment does not solely depend on \\skill level, but on the enthusiasm and engagement in the \\activity. \\
\textit{Con Rationale:} The phrase "enjoy yourself" implies a level of\\ satisfaction and contentment, which contradicts the notion of \\James not being good at surfing, as his lack of skill is likely to \\cause frustration rather than enjoyment. Therefore, it is \\implausible that James would consistently enjoy himself given \\his admitted lack of proficiency.\\
Human Rating $\Delta$: $-0.6$ \\
LLM Rating $\Delta$ (OpenAI): $+0.75$\\
LLM Rating $\Delta$ (Non-OpenAI): $+0.11$
} \\
\midrule

\specialcell{Human Rating $\downarrow$ \\ \\ LLM Rating $\downarrow$} & \specialcellleft{\textit{Question}: Aside from water and nourishment what does your \\dog need? \\
\textit{Choice}: lots of attention \textit{(gold label)}  \\ \textit{Pro Rationale:} Dogs are social animals that thrive on interaction \\and companionship, making lots of attention essential for their \\emotional well-being and development. By providing \\attention, owners can strengthen their bond with their dog, \\enhance socialization skills, and reduce the risk of behavioral \\issues stemming from loneliness or neglect.\\
\textit{Con Rationale:} While dogs may enjoy attention, it is not a\\ fundamental need comparable to shelter or regular exercise, \\making "lots of attention" an implausible equivalent to basic \\survival necessities.\\
Human Rating $\Delta$: $-0.4$  \\
LLM Rating $\Delta$ (OpenAI): $-1$ \\
LLM Rating $\Delta$ (Non-OpenAI): $-0.33$
} \\
\bottomrule
\end{tabularx}
\caption{(Continued from \autoref{pro_con_examples})Examples from \textsc{SIQA} and \textsc{CQA} showing mean plausibility rating changes for different agents when \texttt{PRO Rationales} and \texttt{CON Rationales} are shown together.}\label{pro_con_examples_contd}
\end{table*}

\begin{figure*}[ht]
    \centering
    \fbox{\includegraphics[width=1\linewidth]{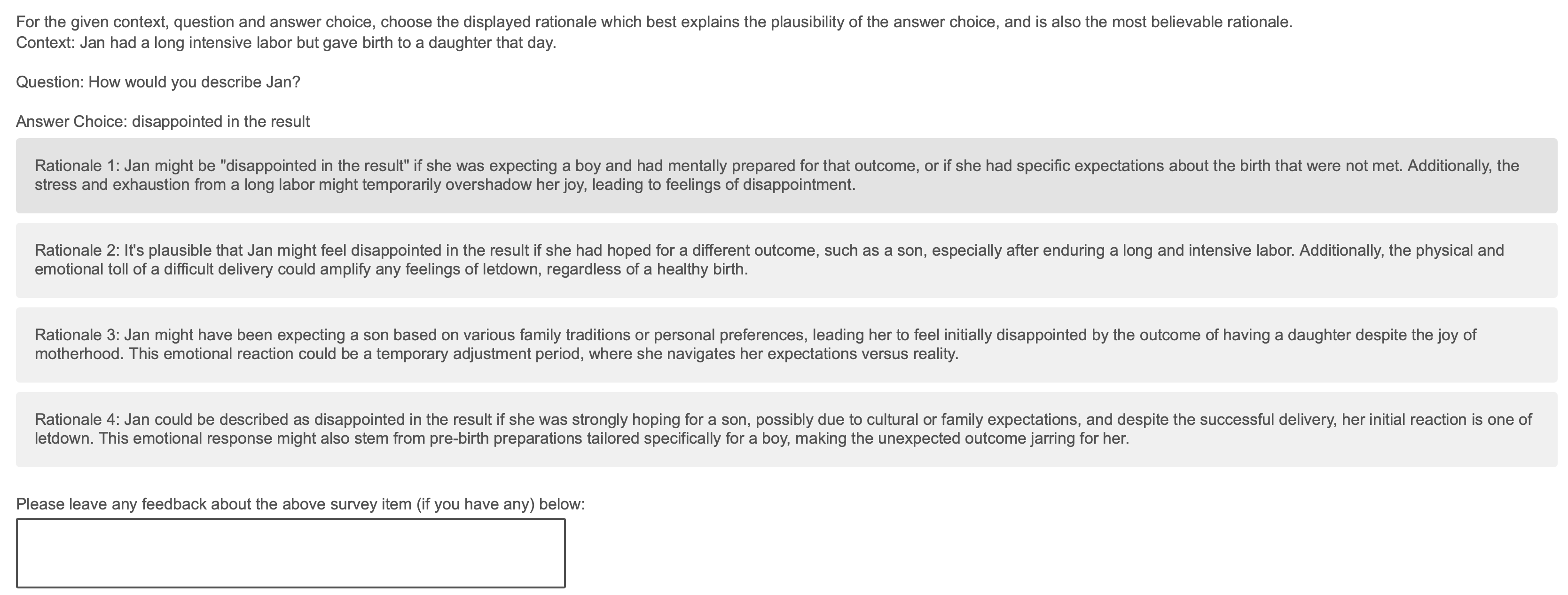}}
    \caption{\label{preference_interface} An example of the interface that annotators used while choosing the best rationale for an answer choice as described in \cref{sec:preference}. Example taken from \textsc{SIQA}.}
\end{figure*}

\begin{figure*}[ht]
    \centering
    \fbox{\includegraphics[width=1\linewidth]{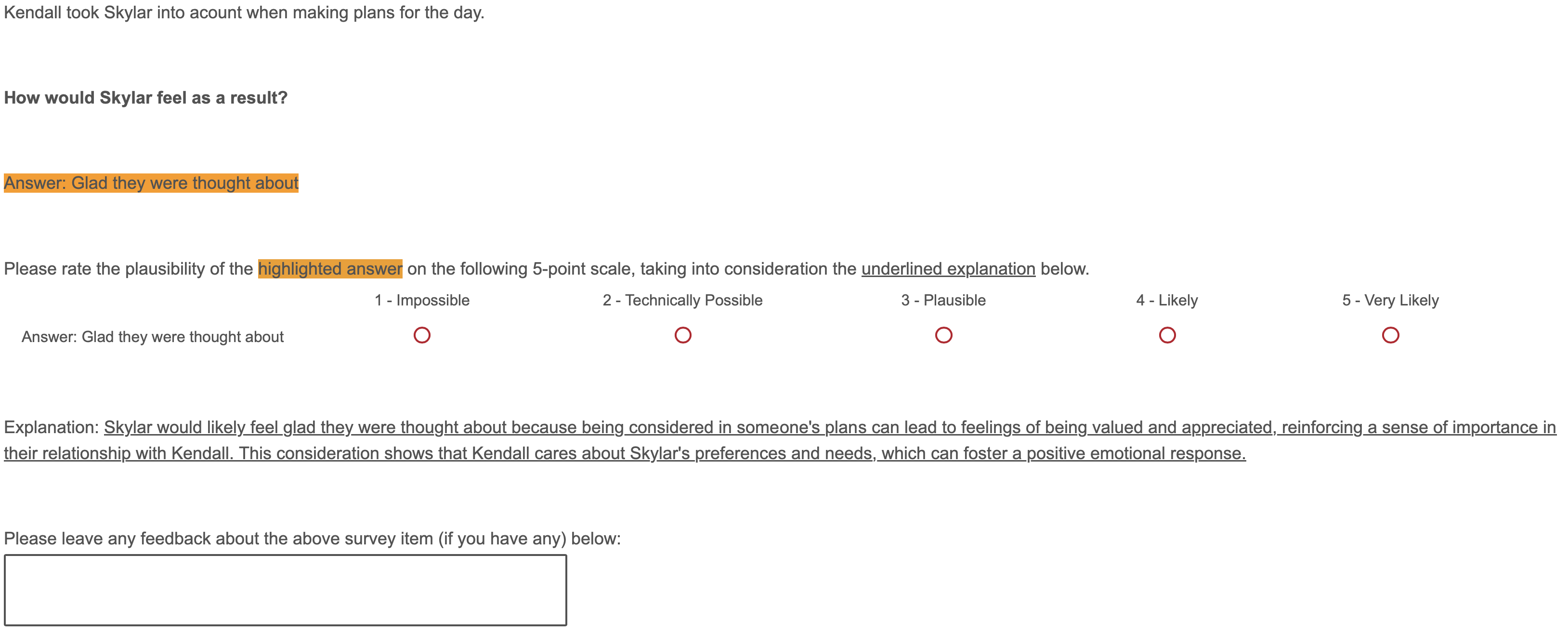}}
    \caption{\label{siqa_pro_interface} An example of the interface that annotators used while giving plausibility ratings to answer choices with a \texttt{PRO} rationale as described in \cref{sec:human_ratings}. Example taken from \textsc{SIQA}.}
\end{figure*}

\begin{figure*}[ht]
    \centering
    \fbox{\includegraphics[width=1\linewidth]{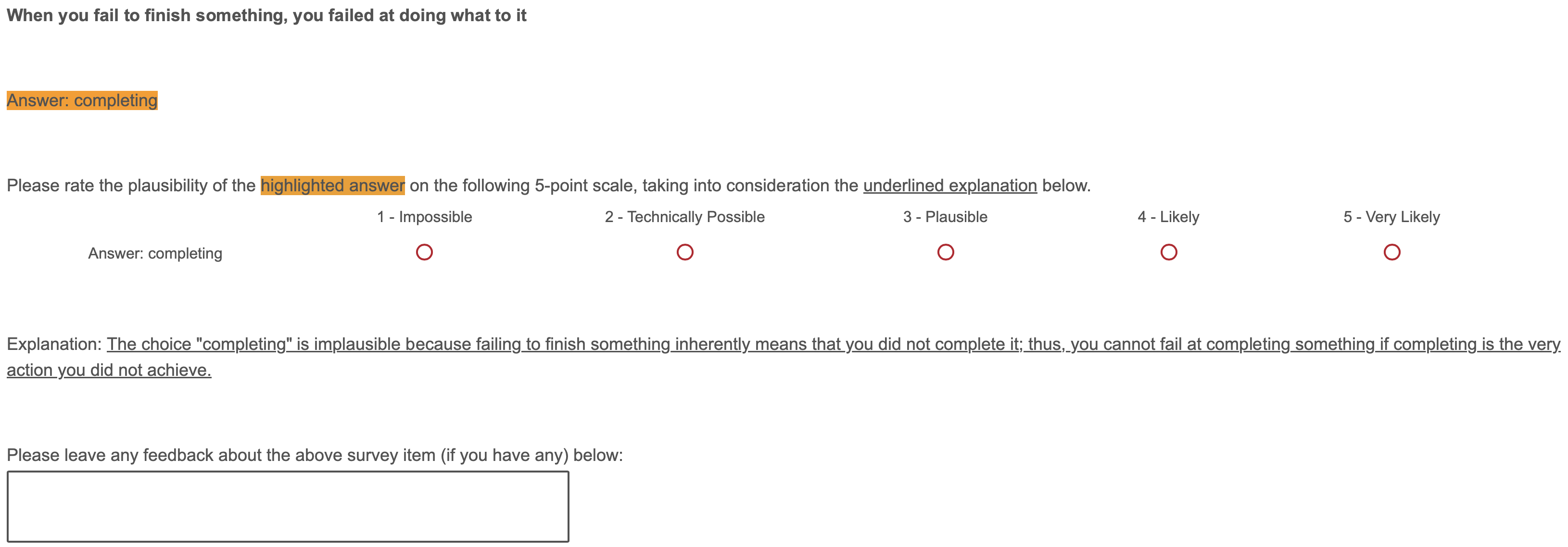}}
    \caption{\label{cqa_con_interface} An example of the interface that annotators used while giving plausibility ratings to answer choices with a \texttt{CON} rationale as described in \cref{sec:human_ratings}. Example taken from \textsc{CQA}.}
\end{figure*}

\begin{figure*}[ht]
    \centering
    \fbox{\includegraphics[width=1\linewidth]{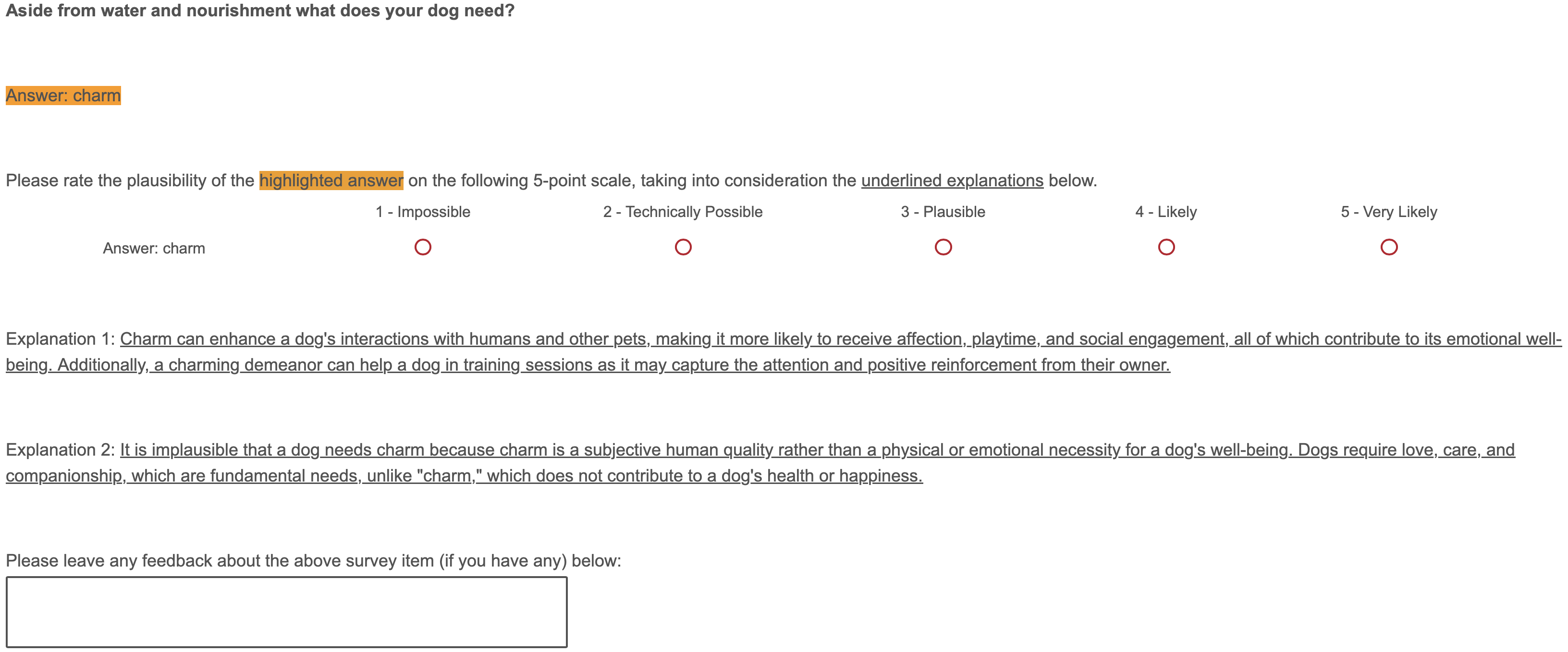}}
    \caption{\label{cqa_pro_con_interface} An example of the interface that annotators used while giving plausibility ratings to answer choices with both \texttt{PRO and CON} rationales as described in \cref{sec:human_ratings}. Example taken from \textsc{CQA}.}
\end{figure*}
\end{document}